%% file: main.tex
\documentclass[10pt,twocolumn,letterpaper]{article}

\usepackage{cvpr}
\input{preamble}

\def\paperID{*****}
\def\confName{CVPR}
\def\confYear{2026}

\title{WAVE-Stereo: Warp-Aligned Volume Encoding for Stereo Matching}

\author{Zehan Liu\\
School of Electronics and Control\\
Engineering\\
Chang'an University\\
Xi'an, China\\
{\tt\small zehanliu@chd.edu.cn}
\and
Yage He\\
School of Electronics and Control\\
Engineering\\
Chang'an University\\
Xi'an, China\\
{\tt\small heyage@chd.edu.cn}
\and
Xianwu Gong$^*$\\
School of Electronics and Control\\
Engineering\\
Chang'an University\\
Xi'an, China\\
{\tt\small xwgong@chd.edu.cn}
}

\begin{document}
\maketitle

\input{sec/0_abstract}
\input{sec/1_intro}
\input{sec/2_related_work}
\input{sec/3_method}
\input{sec/4_experiments}
\input{sec/5_conclusion}
\input{sec/6_ack}

{
    \small
    \bibliographystyle{ieeenat_fullname}
    \bibliography{main}
}

\end{document}

%% file: preamble.tex
\usepackage{indentfirst}
\usepackage{graphicx}
\usepackage{amsmath}
\usepackage{amssymb}
\usepackage{booktabs}
\usepackage{multirow}
\usepackage{makecell}
\usepackage[table]{xcolor}
\usepackage{subcaption}

\definecolor{cvprblue}{rgb}{0.21,0.49,0.74}
\usepackage[pagebackref,breaklinks,colorlinks,allcolors=cvprblue]{hyperref}

\usepackage[capitalize]{cleveref}
\crefname{section}{Sec.}{Secs.}
\Crefname{section}{Section}{Sections}
\Crefname{table}{Table}{Tables}
\crefname{table}{Tab.}{Tabs.}
\Crefname{figure}{Figure}{Figures}
\crefname{figure}{Fig.}{Figs.}



%% file: sec/0_abstract.tex
\begin{abstract}
Existing iterative stereo matching methods primarily adopt two types of correspondence representation: explicit matching search via correlation volumes and local residual refinement via warped features, yet the two remain separately modeled. We propose WAVE-Stereo, built on a core insight: correlation volumes and feature warping provide complementary matching cues. \textbf{GeoWarp Correspondence Encoder (GWCE)} encodes matching search, residual alignment, and disparity prior in parallel at the ConvGRU input. To mitigate matching degradation in textureless regions, we propose \textbf{Periodic Global Context Propagation (PGCP)}, which propagates global spatial information in a periodic manner. On five real-world benchmarks---Middlebury, ETH3D, KITTI 2012, KITTI 2015, and Booster---WAVE-Stereo achieves competitive zero-shot generalization accuracy without any external foundation model prior, achieving 3.18\% D1-all on KITTI 2015, 4.42\% Bad-2.0 on Booster, and 66ms real-time inference, striking a favorable balance between accuracy and efficiency. Our code is available at \url{https://github.com/yamanoko-do/WAVE-Stereo}.
\end{abstract}

%% file: sec/1_intro.tex
\section{Introduction}
\label{sec:intro}

\begin{figure}[t]
\centering
\vspace{-10pt}
\includegraphics[width=0.95\columnwidth]{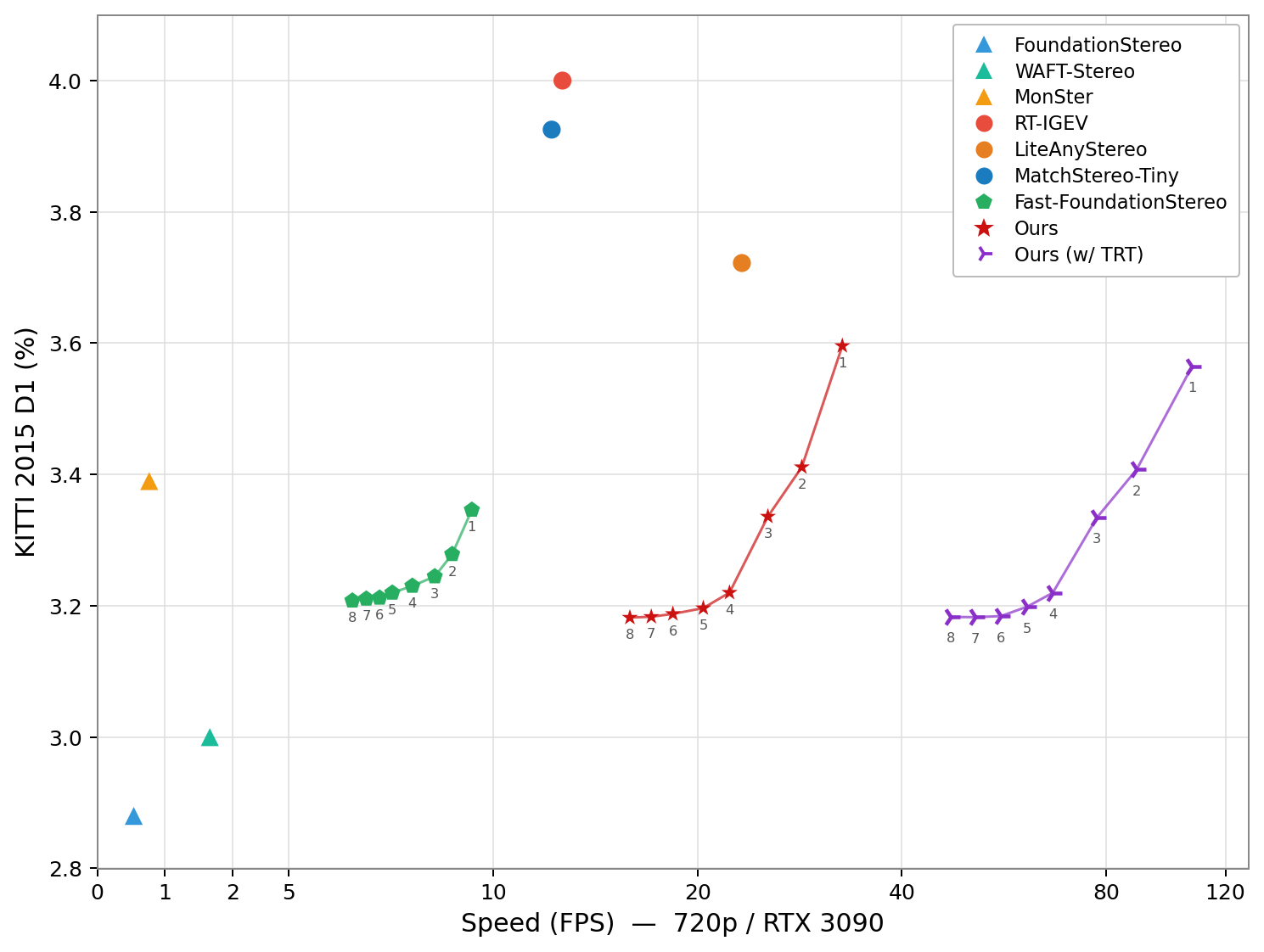}
\vspace{-0.15in}
\caption{Speed-accuracy trade-off on KITTI 2015. WAVE-Stereo achieves competitive generalization performance with significantly faster inference.}
\label{fig:scatter}
\vspace{-10pt}
\end{figure}

Stereo matching aims to estimate dense disparity maps from rectified binocular image pairs and reconstruct 3D scene geometry, serving as a key visual perception task in autonomous driving, robotic navigation, and augmented reality systems. The core challenge lies in the matching ambiguity present in textureless regions, repetitive structures, specular highlights, transparent surfaces, and occluded areas.

Over the past decade, deep learning has steadily advanced stereo matching accuracy. Early methods modeled the problem as cost volume filtering: constructing 3D or 4D cost volumes from left-right image features, then applying 2D or 3D convolutional networks for regularization before outputting disparity maps. However, 3D convolutions incur substantial computational and memory costs, limiting both spatial resolution and computational efficiency.

RAFT-Stereo~\cite{raft-stereo} introduced the iterative optimization paradigm from optical flow into stereo matching, using ConvGRU to sequentially update the disparity field. At each iteration, local matching costs are retrieved from a pre-computed all-pairs correlation pyramid. This method uses only 2D convolutions and lightweight correlation computation, achieving leading generalization performance on multiple real-world benchmarks. IGEV-Stereo~\cite{igev} further introduced a geometry encoding volume: a lightweight 3D CNN regularization is applied to the initial correlation volume, constructing a multi-level pyramid that enables each ConvGRU update to retrieve features encoding both scene geometry and local matching details. This design significantly reduces the number of iterations required for convergence.

However, existing iterative stereo matching methods exhibit a clear paradigm split in correspondence representation. One class~\cite{igev} relies on correlation volumes for explicit matching search, preserving matching ambiguity and search space among disparity candidates through pre-computed matching costs. The other class~\cite{waft-stereo}, inspired by~\cite{waft}, employs feature warping---using the current disparity estimate to align right-image features to the left-image coordinate system and performing iterative refinement on cross-view residuals. The former preserves candidate matching information, while the latter leverages cross-view residuals for local optimization. Yet, existing methods typically adopt only one of these paradigms: correlation-based methods lack cross-view alignment information under the current estimate, while warping-based methods discard the ambiguity information in matching costs, leading to erroneous alignment in ill-posed regions. WAFT-Stereo~\cite{waft-stereo} first demonstrated that a pure warping paradigm can completely eliminate dependence on cost volumes, but its feature encoder employs Depth-AnythingV2-Large~\cite{depthanythingv2}, incurring substantial inference overhead. To date, no work has unified these two types of correspondence representations.

\begin{figure}[t]
\centering
\includegraphics[width=\columnwidth]{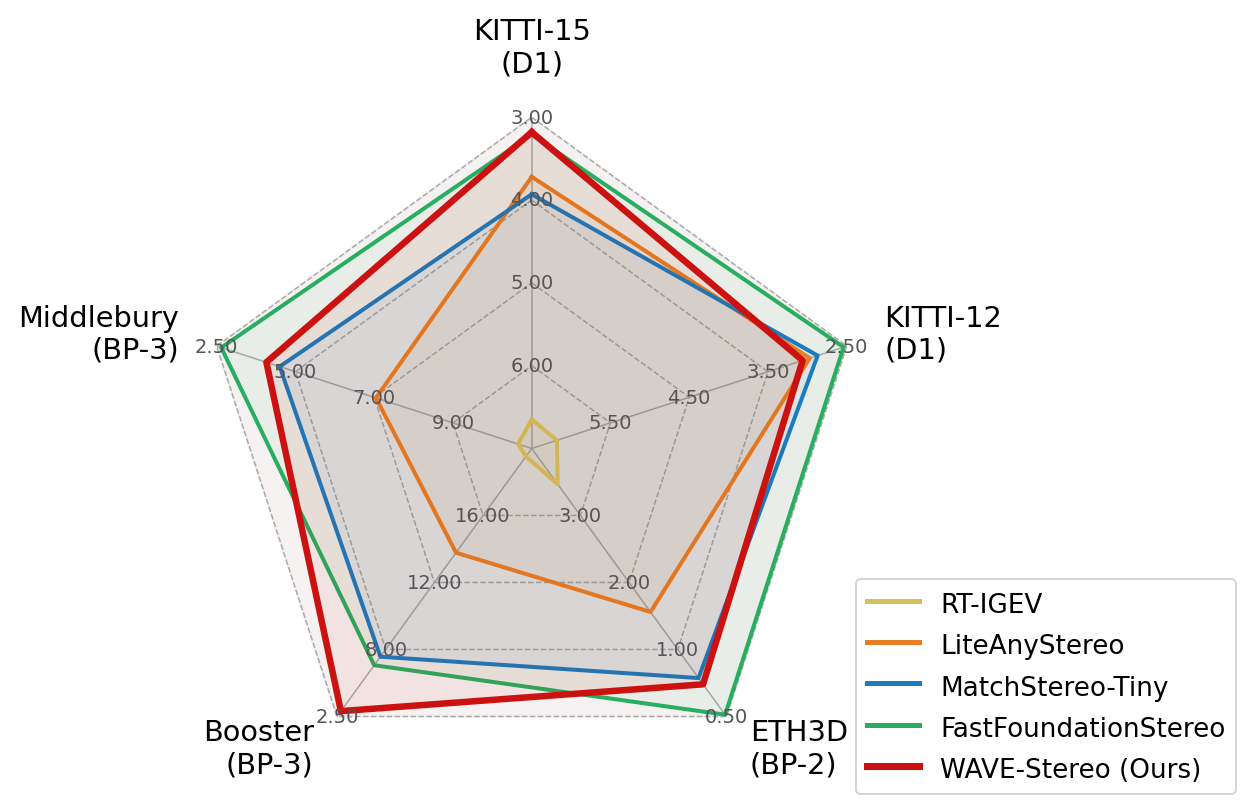}
\caption{Zero-shot evaluation on five real-world benchmarks. Our method demonstrates competitive zero-shot generalization accuracy.}
\label{fig:radar}
\end{figure}

Furthermore, the iterative update of ConvGRU is inherently limited to a local context window. In textureless walls, repetitive patterns, or transparent surfaces, local matching is ambiguous without global spatial context. Recent methods mitigate this by introducing ViT backbones or monocular depth foundation models, but these models incur inference times of hundreds of milliseconds, precluding real-time deployment.

We propose WAVE-Stereo (Warp-Aligned Volume Encoding Stereo), built on the following observation: correlation volumes and feature warping provide two complementary types of correspondence representation. The former preserves matching information and its ambiguity among disparity candidates through explicit matching costs; the latter aligns right-image features to the left-image coordinate system based on the current disparity estimate, transforming stereo matching into a local residual correction problem. Through GWCE, we unify both representations at the ConvGRU input, enabling each iteration to simultaneously leverage matching candidate information and cross-view alignment signals. To periodically inject global spatial constraints into the iterative process with minimal overhead, we design PGCP.

Our contributions are summarized as follows:
\begin{itemize}
\item We propose WAVE-Stereo, a lightweight iterative stereo matching framework that achieves good zero-shot generalization while maintaining real-time inference, without requiring any external foundation model.

\item We propose \textbf{GeoWarp Correspondence Encoder (GWCE)}, which addresses the limitation of existing methods that employ only a single correspondence representation (correlation search or feature warping). GWCE encodes three complementary types of information---correspondence search, residual alignment, and geometric prior---at the ConvGRU input, achieving unified modeling of matching search and residual optimization.

\item We propose \textbf{Periodic Global Context Propagation (PGCP)}, designed to address the inability of ConvGRU's local iterations to propagate long-range matching constraints. By periodically injecting global spatial constraints during iteration, PGCP introduces global context for ambiguous regions, improving disparity estimation in textureless areas.

\item We validate a favorable accuracy-efficiency balance across multiple benchmarks: competitive in-domain accuracy on SceneFlow, and competitive zero-shot generalization on Middlebury, ETH3D, KITTI, and Booster with 66ms inference speed, without any external foundation model.
\end{itemize}

\begin{figure*}[t]
\centering
\includegraphics[width=\textwidth]{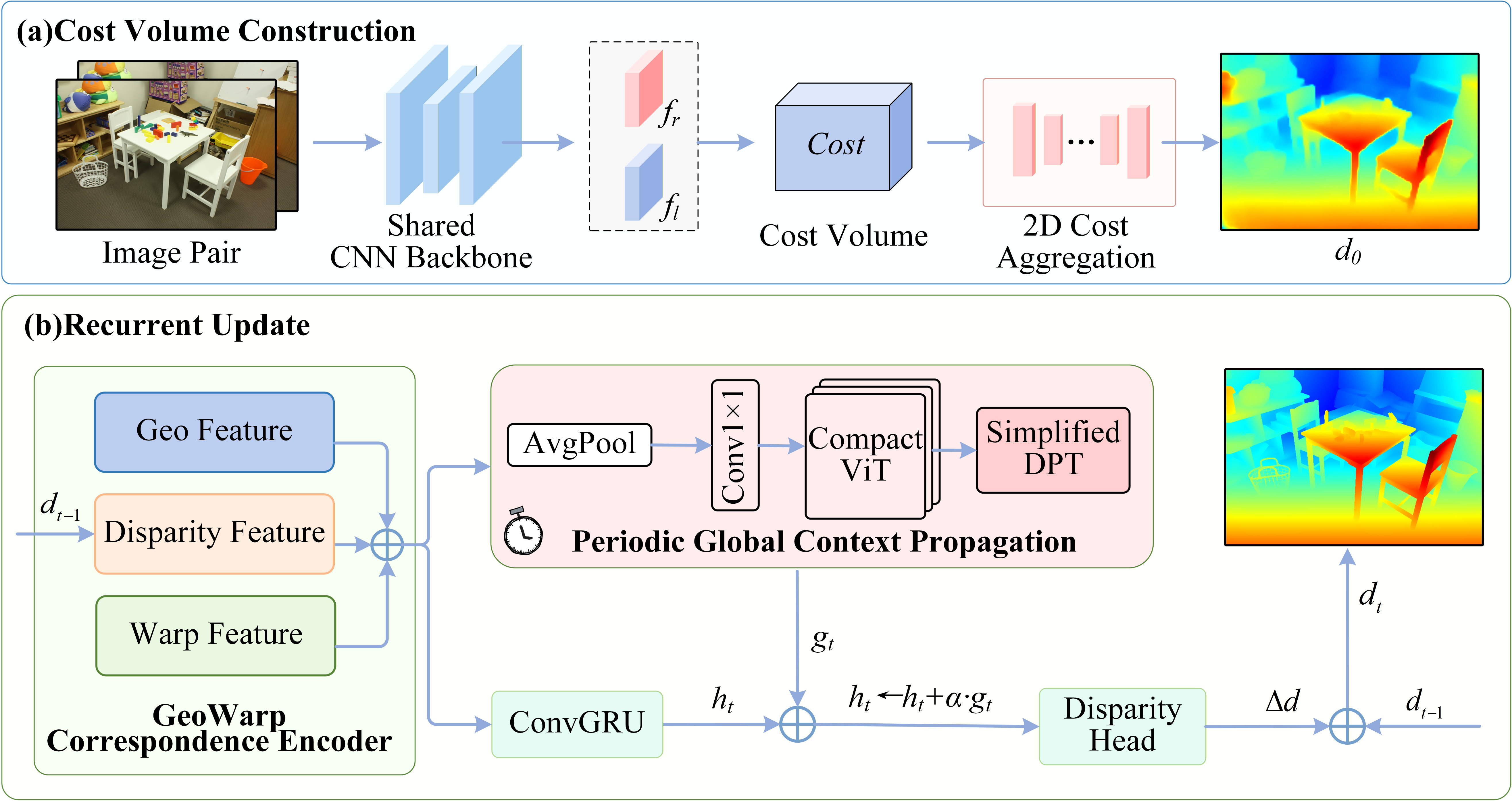}
\caption{Overall architecture of WAVE-Stereo. The pipeline consists of two key components: GWCE unifies correlation search, residual alignment, and geometric prior in parallel at the ConvGRU input; PGCP periodically injects global spatial constraints at 1/32 resolution.}
\label{fig:pipeline}
\end{figure*}

%% file: sec/2_related_work.tex
\section{Related Work}
\label{sec:related}

\begin{figure*}[t]
\centering
\includegraphics[width=\textwidth]{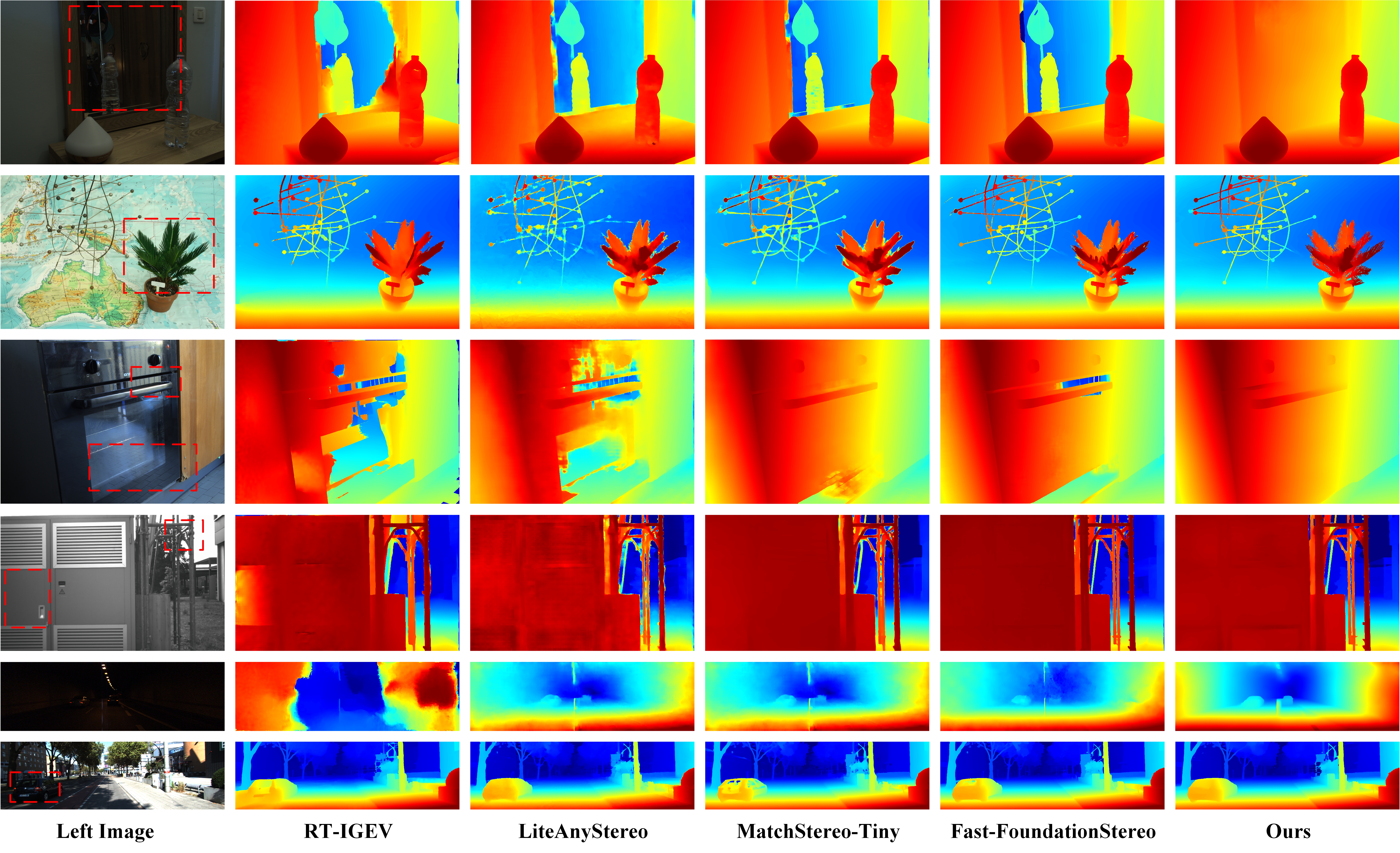}
\caption{Visualized qualitative comparisons for real-time disparity estimation methods are displayed on Middlebury, ETH3D, Booster and KITTI-2015. All predictions stem from zero-shot evaluation.}
\label{fig:zeroshot_compare}
\end{figure*}

\subsection{Iterative Stereo Matching}
\label{sec:iterative}

The core idea of iterative stereo matching originates from RAFT~\cite{raft}: pre-compute an all-pairs feature correlation pyramid, retrieve local matching costs at each iteration, and have a ConvGRU predict residual flow to progressively refine the disparity field. RAFT-Stereo~\cite{raft-stereo} adapted this paradigm to stereo matching, replacing 3D convolutional cost aggregation with cross-connected multi-resolution ConvGRUs---multiple GRUs independently maintain hidden states at 1/16 and 1/8 resolution, exchanging information across adjacent levels through upsampling and downsampling, demonstrating the superior cross-domain generalization capability of pure 2D convolutional iterative architectures.

However, local retrieval from all-pairs correlations lacks non-local geometric information, leading to ambiguity in ill-posed regions. IGEV-Stereo~\cite{igev} proposed a Combined Geometry Encoding Volume: first computing a group-wise correlation volume, applying a three-level 3D hourglass network for cost aggregation (with channel expansion from 16 to 48 guided by left-image feature channel excitation), then combining it with the all-pairs correlation pyramid. The initial disparity regressed from this volume is far superior to zero initialization, dramatically accelerating ConvGRU convergence. IGEV++~\cite{igevpp} further constructs geometry encoding volumes at small, medium, and large disparity ranges, covering broader disparity ranges through adaptive patch matching. While these methods achieve leading accuracy, the 3D convolutional regularization and multi-range volumes push inference time to hundreds of milliseconds.

In contrast to the cost-volume-based approach, the WAFT series~\cite{waft,waft-stereo} fundamentally restructured the iterative matching architecture. WAFT~\cite{waft} demonstrated in optical flow that iterative matching can be achieved solely through high-resolution feature warping without explicitly constructing cost volumes. WAFT-Stereo~\cite{waft-stereo} extended this idea to stereo matching, replacing traditional correlation computation and cost volume construction with feature warping, and providing more stable initial estimates through disparity classification, achieving high accuracy with improved efficiency. However, the pure warping paradigm completely discards explicit matching candidates and ambiguity distribution information, lacking the multi-candidate search space preserved by correlation-based methods; moreover, its performance depends on large-scale pretrained depth models as backbones, hindering lightweight deployment.

WAVE-Stereo follows the ConvGRU iterative optimization paradigm pioneered by RAFT-Stereo~\cite{raft-stereo} and uses the geometry encoding volume proposed by IGEV~\cite{igev} for matching cost retrieval. Unlike existing methods that employ only a single correspondence representation, GWCE unifies correspondence search (correlation retrieval) and residual alignment (feature warping), enabling each ConvGRU update to simultaneously receive matching candidate information and cross-view alignment signals. Furthermore, PGCP supplements the limited receptive field of local iterative updates through periodic global spatial constraint injection.

\subsection{Lightweight Stereo Matching}
\label{sec:lightweight}

Real-time performance is a critical requirement for practical stereo matching deployment, yet the computational and memory overhead of traditional 3D cost volume aggregation severely constrains inference speed. Existing lightweight methods mainly pursue efficiency along three directions.

The first category reduces cost volume complexity while retaining 3D convolutions. StereoNet~\cite{stereonet} constructs cost volumes on low-resolution features and recovers high-frequency details through color-image-guided hierarchical refinement. BGNet~\cite{bgnet} proposes edge-aware cost volume upsampling, confining expensive operations such as 3D convolutions to low resolutions. CoEx~\cite{bangunharcana2021correlate} improves aggregation efficiency through guided cost volume excitation, while Fast-ACVNet~\cite{xu2023accurate} selectively processes high-resolution matching with sparse attention. Although these methods reduce absolute computation, the inherent overhead of 3D convolutions still limits further efficiency gains.

The second category replaces 3D convolutions with pure 2D operations. AANet~\cite{xu2020aanet} adopts deformable 2D convolutions for adaptive cost aggregation; MobileStereoNet-2D~\cite{shamsafar2022mobilestereonet} builds lightweight backbones with MobileNet modules. LightStereo~\cite{lightstereo} further proposes a key insight: through substantial channel expansion (Channel Boost) in 2D cost aggregation, one can match the accuracy of 3D/4D cost volumes without relying on 3D convolutions. Its core module employs MobileNetV2~\cite{mobilenetv2} inverted residual blocks that temporarily expand input channels, perform spatial propagation via depthwise separable convolutions, and project back to the original channel count, combined with a three-level 2D UNet and multi-scale stripe attention, achieving strong performance in both speed and accuracy. However, pure 2D methods generally face limited cross-domain generalization---most models are optimized for specific domains (particularly the KITTI online benchmark) and exhibit restricted generalization on diverse unseen scenes.

The third category adopts hybrid 3D-2D aggregation. LiteAnyStereo~\cite{liteanystereo} retains a small 3D component to maintain disparity awareness while using ConvNeXt-style 2D layers for spatial aggregation, achieving strong zero-shot generalization through a three-stage training strategy (supervised pretraining, self-distillation, pseudo-label distillation). This method demonstrates that through careful training strategy design, lightweight models can also achieve generalization comparable to foundation-model-enhanced methods.

WAVE-Stereo's cost aggregation module follows LightStereo's~\cite{lightstereo} 2D aggregation design: MobileNetV2~\cite{mobilenetv2} inverted residual blocks for channel expansion and a three-level UNet for multi-scale feature fusion. Building on the initial disparity from 2D aggregation, WAVE-Stereo further applies iterative refinement: GWCE unifies correlation retrieval and feature warping as motion encoding, and PGCP periodically injects global spatial constraints. This iterative refinement framework enables progressive refinement of the initial estimate from the lightweight aggregation module, significantly improving zero-shot generalization without increasing cost volume construction overhead.

\subsection{Zero-Shot Generalization in Stereo Matching}
\label{sec:zeroshot_rw}

Maintaining accurate estimation on completely unseen target domains is a critical prerequisite for practical stereo deployment. Traditional methods fine-tuned on specific datasets (e.g., KITTI) exhibit limited cross-domain generalization. Recent research on zero-shot generalization has primarily proceeded along two directions: introducing large-scale pretrained monocular depth foundation models (VFM) into the stereo pipeline to leverage their rich scene priors, or improving generalization through training strategies such as distillation and multi-stage training.

One approach directly introduces large-scale pretrained monocular depth foundation models into the stereo pipeline, leveraging their rich scene priors to enhance generalization. DepthAnythingV2~\cite{depthanythingv2} acquired strong scene geometry priors through large-scale monocular training. FoundationStereo~\cite{foundationstereo} injects the monocular prior of DepthAnythingV2~\cite{depthanythingv2} into the stereo pipeline through side-tuning adapters, combined with axial-planar convolutions and a disparity Transformer, achieving excellent zero-shot generalization across multiple real-world benchmarks, but with inference latency as high as 496ms. MonSter~\cite{monster} jointly refines monocular and multi-view depth, achieving bidirectional complementarity through stereo-guided alignment and monocular-guided refinement, with 336ms inference. BridgeDepth~\cite{bridgedepth} unifies monocular and stereo inference through latent-space cross-attention, significantly improving efficiency while maintaining fast inference. The core cost of these methods is that inference still requires running the entire monocular depth backbone, consuming a large proportion of the computational budget and making edge deployment difficult.

The second approach improves generalization through distillation and multi-stage training strategies, rather than introducing external models at inference time. Fast-FoundationStereo~\cite{fastfoundationstereo} uses FoundationStereo as a teacher model, compressing the VFM's generalization capability into a smaller model through knowledge distillation, block-wise architecture search, and structured pruning, reducing inference time to 188ms. LiteAnyStereo~\cite{liteanystereo} completely bypasses VFMs: on a lightweight ConvNeXt-style 2D aggregation architecture, it achieves competitive zero-shot generalization through three-stage training---supervised pretraining, self-distillation, and pseudo-label distillation---without relying on any external model. Both works demonstrate from different angles that carefully designed training strategies can preserve or even approach the generalization level of VFM-based methods without carrying a monocular VFM at inference time.

WAVE-Stereo combines the insights of both directions: on the data side, it trains on large-scale data similar to FoundationStereo~\cite{foundationstereo}, ensuring exposure to sufficiently diverse scene distributions during training; on the model architecture side, GWCE unifies matching search and residual alignment, and PGCP periodically injects global spatial constraints, enabling the model to more fully exploit training signals at a given data scale. Compared to VFM-dependent methods, WAVE-Stereo achieves competitive generalization results on multiple real-world benchmarks without requiring an external monocular depth backbone. Compared to LiteAnyStereo~\cite{liteanystereo}'s multi-stage training, WAVE-Stereo achieves strong zero-shot generalization with only single-stage supervised training, through carefully designed correspondence representations and sufficiently diverse training data.

\begin{figure}[t]
\centering
\includegraphics[width=\columnwidth]{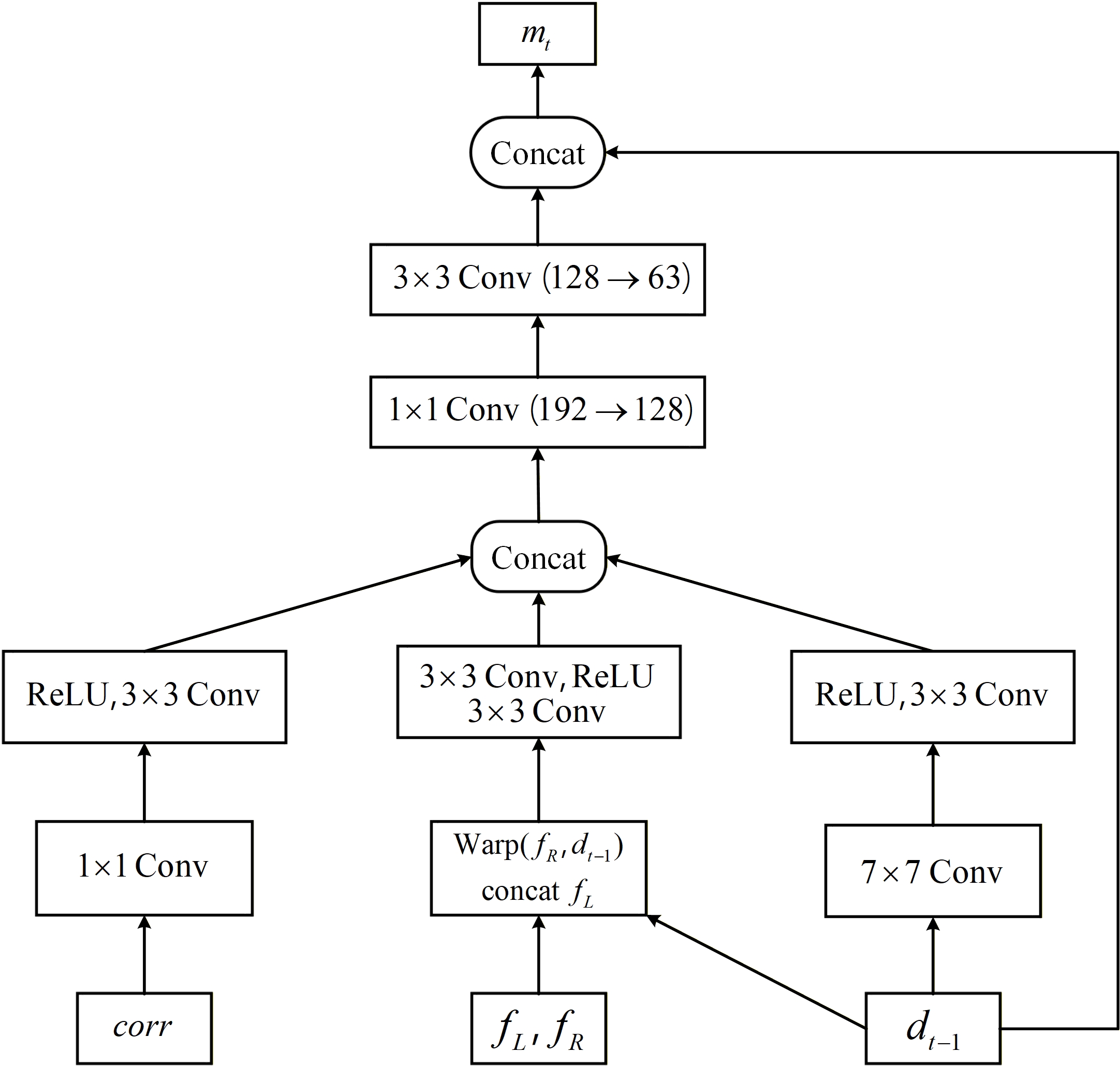}
\caption{Detailed architecture of the GeoWarp Correspondence Encoder (GWCE). Three parallel branches---correlation search, disparity prior, and cross-view warping alignment---encode complementary matching signals, which are fused to produce motion features $\mathbf{m}_t$ for the ConvGRU.}
\label{fig:gwce}
\end{figure}

%% file: sec/3_method.tex
\section{Method}
\label{sec:method}

The overall network architecture is illustrated in Fig.~\ref{fig:pipeline}. We first present the two core modules, GWCE and PGCP, followed by the overall training and inference pipeline.

\subsection{GeoWarp Correspondence Encoder}
\label{sec:gwce}

GWCE is based on the following design principle: geometric correlation and feature warping provide fundamentally different matching cues. The former retrieves local matching costs from the cost volume, preserving explicit matching differences and ambiguity information among disparity candidates---in textureless or repetitive regions, multiple disparity candidates may exhibit similar matching scores, and this ambiguity is itself an informative matching signal. The latter warps right-image features to the left-image coordinate system based on the current disparity estimate, enabling the network to directly observe aligned cross-view feature differences, transforming correspondence estimation into a local residual correction problem that is more amenable to iterative refinement. These two provide matching search information and residual alignment information, respectively, and are complementary in nature.

In iterative stereo matching, the motion encoder is responsible for encoding multiple matching signals into the ConvGRU's motion input. Existing methods exhibit a paradigm split in correspondence representation: RAFT-Stereo~\cite{raft-stereo} and IGEV-Stereo~\cite{igev} use only correlation features for matching search, while WAFT-Stereo~\cite{waft-stereo} uses only warped features for residual optimization. GWCE unifies both, providing the GRU with richer matching cues.

The encoder receives three types of input: correlation features $\mathbf{c}$ retrieved from the geometry encoding volume, the current disparity estimate $d$, and left-right image features $f_L, f_R$. These three inputs carry information from different stages of correspondence optimization and are encoded in parallel through three independent branches before fusion.

\textbf{Correlation branch.} This branch retrieves local correlation features from the geometry encoding volume to provide explicit matching search information. The correlation features $\mathbf{c}$ are processed by $\text{Enc}_c$, where a $1\times1$ convolution compresses the channel dimension and unifies feature representations across scales, followed by a $3\times3$ convolution that incorporates local spatial context to produce the correspondence features $\mathbf{x}_c$. Unlike directly regressing a single match, the correlation representation preserves the distribution of matching responses across disparity candidates, enabling the network to leverage relative confidence relationships among them. In textureless or repetitive regions, multiple disparity candidates may produce similar matching responses, and this ambiguity information provides a basis for subsequent ConvGRU updates:
\begin{equation}
\mathbf{x}_c = \text{Enc}_c(\mathbf{c})
\end{equation}

\textbf{Disparity prior branch.} This branch explicitly injects the current disparity estimate into the motion encoding. The current disparity $d$ is processed by encoder $\text{Enc}_d$: a $7\times7$ large-kernel convolution captures sufficiently wide spatial context, enabling the branch to perceive spatial distribution patterns of the current estimate, followed by ReLU and a $3\times3$ convolution. This geometric prior provides the iterative update with current geometric state information:
\begin{equation}
\mathbf{x}_d = \text{Enc}_d(d)
\end{equation}

\textbf{Cross-view warping alignment branch.} This branch transforms correspondence estimation from matching search into local residual correction. First, $f_R$ is warped to the left-image coordinate system via grid sampling using the current disparity $d$ to obtain $\tilde{f}_R$, then concatenated with $f_L$ along the channel dimension and processed by encoder $\text{Enc}_w$ (two $3\times3$ convolutions interleaved with ReLU). Unlike the correlation branch, the warping alignment branch directly observes aligned cross-view feature differences---when the disparity estimate is accurate, $\tilde{f}_R$ and $f_L$ are highly consistent and differences mainly stem from noise; when bias exists, differences are significantly amplified at local edges and textures. The network leverages these cross-view differences to estimate the disparity residual, avoiding the computational cost of explicit search in the candidate disparity space:
\begin{equation}
\tilde{f}_R = \text{Warp}(f_R, d)
\end{equation}
\begin{equation}
\mathbf{x}_w = \text{Enc}_w([f_L, \tilde{f}_R])
\end{equation}
where $\text{Warp}$ denotes the warping operation using bilinear grid sampling, and $[\cdot, \cdot]$ denotes concatenation along the channel dimension.

\textbf{Fusion.} The three branch outputs are first fused through $\text{Fusion}$, then concatenated with the current disparity $d$ to form the motion features $\mathbf{m}_t$, which serve as input to the ConvGRU. The three branches respectively provide matching candidate information, cross-view alignment information, and current geometric state information:
\begin{equation}
\mathbf{m}_t = [\text{Fusion}([\mathbf{x}_c, \mathbf{x}_d, \mathbf{x}_w]), d]
\end{equation}

\subsection{Periodic Global Context Propagation}
\label{sec:pgcp}

The local iterative update of ConvGRU can only propagate short-range matching information, with a limited receptive field. In textureless and repetitive regions, local matching signals are highly ambiguous, and matching ambiguity cannot be resolved through neighborhood context alone---long-range spatial constraints are needed to determine correct correspondences. However, running global self-attention at every iteration is prohibitively expensive, and not every iteration requires global information injection.

The design motivation of PGCP is not to enhance feature expressiveness, but to periodically re-inject global spatial constraints during the iterative process. When local GRU updates drift in ambiguous regions, the global context provides global spatial constraints for the local update. Specifically, PGCP runs lightweight self-attention at extremely low resolution (1/32) only at fixed-interval iterations, keeping the additional computational overhead at a very low level.

The motion features $\mathbf{m}_t$ are pooled by $8\times$ average pooling ($H/4 \to H/32$), projected to a Transformer hidden dimension $D_{tf}=128$ via a $1\times1$ convolution, and flattened into $T = (H/32) \times (W/32)$ tokens. Three Transformer layers, each with 4 attention heads and MLP expansion ratio 3, perform self-attention. The outputs of the first and third layers are reshaped back to spatial dimensions, forming 2 feature scales, which are fed into a simplified DPT module.

The DPT module fuses multi-scale features in a bottom-up manner: starting from the deepest level ($H/32$), each step passes through a DPT fusion block (deep residual convolution + bilinear $2\times$ upsampling), adds the result to the higher-resolution feature, and ultimately returns to $H/4$ resolution. The output is added to the GRU hidden state via a learnable gating parameter $\alpha$:
\begin{equation}
h_t = h_t + \alpha \cdot \mathbf{g}_t
\end{equation}
where $\mathbf{g}_t$ is the global feature output from DPT, and $\alpha$ is a learnable scalar initialized to $-0.1$.

\subsection{Overall Architecture}
\label{sec:overall}

We now present the overall architecture of WAVE-Stereo. Given a rectified binocular image pair $\mathbf{I}_L, \mathbf{I}_R \in \mathbb{R}^{3\times H\times W}$, a shared-weight MobileNetV2~\cite{mobilenetv2} lightweight backbone first extracts multi-scale features, which are fused top-down through an FPN with skip connections to produce features at four scales. The 1/4-resolution features $f_L, f_R \in \mathbb{R}^{C_f\times H/4\times W/4}$ ($C_f=24$) are used for all subsequent operations. The maximum disparity $D_{\max}$ corresponds to $D = D_{\max}/4$ disparity candidates at 1/4 resolution.

\textbf{Initial disparity.} We compute all-pairs correlation between $f_L$ and $f_R$ (channel-wise dot product with mean), constructing a 3D cost volume $\mathbf{C} \in \mathbb{R}^{D\times H/4\times W/4}$. A lightweight 2D aggregation module regularizes $\mathbf{C}$. Soft-argmin regression yields the initial disparity $d_0 \in \mathbb{R}^{1\times H/4\times W/4}$:
\begin{equation}
d_0 = \sum_{d=0}^{D-1} d \cdot \text{softmax}(\mathbf{C}(d))
\end{equation}

\textbf{Iterative update.} At iteration $t$, GWCE receives three inputs: (1) local correlation features $\mathbf{c}_t$ retrieved from the geometry encoding volume; (2) the current disparity $d_{t-1}$; (3) warped right-image features $\tilde{f}_R = \text{Warp}(f_R, d_{t-1})$. These three components are encoded through independent convolutional branches, concatenated, and fused into motion features $\mathbf{m}_t \in \mathbb{R}^{64\times H/4\times W/4}$. The motion features are then fed together with context features into the ConvGRU to update the hidden state $h_t$. A disparity head decodes the residual update $\Delta d_t$, and $d_t = d_{t-1} + \Delta d_t$.

\textbf{Periodic global context injection.} Every $K$ iterations, the motion features are pooled by $8\times$ average pooling ($H/4 \to H/32$), passed through a compact ViT and a simplified DPT fusion module to produce global features, which are added to the ConvGRU hidden state via a learnable gating mechanism.

\textbf{Upsampling.} RAFT~\cite{raft}-style convex upsampling (Context Upsampling) recovers the 1/4-resolution disparity to full resolution. The upsampling mask is generated from the GRU's output mask feature map through two-stage deconvolution with progressive upsampling and stem feature fusion.

\textbf{Loss function.} We compute the disparity loss on the initial disparity and all iteration disparities:
\begin{equation}
\mathcal{L}_{\text{stereo}} = \lambda_0 \cdot \text{Smooth}_{L_1}(d_0 - d_{\text{gt}}) + \sum_{i=1}^{N} \gamma^{N-i} \|d_i - d_{\text{gt}}\|_1
\end{equation}
where $d_0$ is the initial disparity from 2D cost aggregation with soft-argmin regression, bilinearly upsampled to full resolution, $\lambda_0=0.5$ is the initial loss weight. $\{d_i\}_{i=1}^N$ are the $N$ iteration disparity predictions (full resolution), and $\gamma=0.9$ is the exponential decay factor assigning higher weight to later iterations.

%% file: sec/4_experiments.tex
\section{Experiments}
\label{sec:experiments}

We evaluate WAVE-Stereo on five public benchmarks: SceneFlow, Middlebury, ETH3D, KITTI, and Booster. We first describe the datasets and implementation details, then present zero-shot generalization and in-domain results, and finally conduct ablation studies to verify the contribution and design choices of each module.

\subsection{Datasets and Evaluation Metrics}
\label{sec:datasets}

We evaluate on five widely-used public datasets.

\textbf{SceneFlow}~\cite{sceneflow} is a synthetic dataset with $960\times540$ resolution, providing 35,454 training pairs and 4,370 test pairs with dense disparity annotations. We use the more challenging Finalpass version (with motion blur and defocus) for both training and testing. The end-point error (EPE) is reported.

\textbf{Middlebury}~\cite{middlebury} is a real-world indoor dataset with $2948\times1988$ resolution, providing two sets of 15 image pairs for training and testing. Images are available at full, half, and quarter resolutions. For evaluation, we use only the half-resolution training pairs to assess cross-domain generalization, reporting Bad-3.0 (percentage of pixels with absolute error $>3$).

\textbf{KITTI 2012/2015}~\cite{kitti2012,kitti2015} are real-world driving datasets at $1242\times375$ resolution. KITTI 2012 contains 194 training and 195 test pairs, while KITTI 2015 provides 200 training and 200 test pairs. Both provide sparse ground-truth disparity from LiDAR. We report D1-all for both KITTI 2012 and KITTI 2015.

\textbf{ETH3D}~\cite{eth3d} contains grayscale stereo images from indoor and outdoor environments, providing 27 pairs with ground truth. We use only the training set for cross-domain generalization evaluation, reporting Bad-2.0.

\textbf{Booster}~\cite{booster} contains extensive scenes with translucent and specular surfaces, evaluating robustness to non-Lambertian surfaces. It provides 228 training and 15 test (without GT) pairs. We use only the training set for cross-domain generalization evaluation, reporting Bad-2.0 and Bad-3.0.

\subsection{Implementation Details}
\label{sec:impl}

WAVE-Stereo is implemented in PyTorch. All experiments use the AdamW~\cite{adamw} optimizer with a OneCycleLR learning rate scheduler, peak learning rate 0.0001, and gradient norm clipping to 0.1. Training uses a batch size of 16 image pairs per GPU on 8 NVIDIA A100 GPUs. The model is jointly trained on 9 public synthetic datasets: SceneFlow~\cite{sceneflow}, FSD~\cite{foundationstereo}, FallingThings~\cite{fallingthings}, IRS~\cite{wang2021irs}, SynClearDepth~\cite{bai2024cleardepth}, VirtualKitti2~\cite{vkitti2}, InStereo2K~\cite{instereo2k}, UnrealStereo4K~\cite{Tosi2021CVPR}, and TartanAir~\cite{tartanair}. Images are randomly cropped to $320\times720$ during training, with data augmentation similar to CREStereo~\cite{crestereo}, including asymmetric color jitter, random erasing, random scaling and stretching, random grayscale, Gaussian blur, and region-based noise injection. The maximum disparity is set to 192. ConvGRU training iterations are set to 12, with 8 iterations during inference by default unless otherwise noted. All speed and memory measurements are performed at $1280\times736$ (720p) resolution on an NVIDIA RTX 3090.

\subsection{Zero-Shot Generalization}
\label{sec:zeroshot}

Table~\ref{tab:zeroshot} presents the comparison on five public real-world datasets. WAVE-Stereo is jointly trained on the mixed dataset and evaluated in a zero-shot manner on Middlebury, ETH3D, KITTI 2015, and Booster---all target-domain data unseen during training. 
\input{table/zero_shot}

Note: All methods are evaluated using their publicly available best weights under the same environment. Middlebury uses half-resolution (H), non-occluded regions; ETH3D uses non-occluded regions; KITTI uses non-occluded regions; Booster uses quarter-resolution (Q). Iterative methods are evaluated with their default inference iteration counts. 
As shown in Table~\ref{tab:zeroshot}, WAVE-Stereo achieves strong performance at 66ms inference speed across five real-world benchmarks. WAVE-Stereo does not rely on any external foundation model---FoundationStereo~\cite{foundationstereo} and MonSter~\cite{monster} both use the monocular prior provided by DepthAnythingV2~\cite{depthanythingv2}, and WAFT-Stereo~\cite{waft-stereo} similarly relies on DepthAnythingV2-Large as its feature backbone. This indicates that feature warping in GWCE helps improve matching performance in transparent and specular regions. On Middlebury and ETH3D's textureless regions, PGCP's periodic global context injection significantly improves disparity continuity.

\input{table/sceneflow}

\subsection{SceneFlow In-Domain Comparison}
\label{sec:sceneflow}

To quantitatively evaluate model performance, we compare WAVE-Stereo with state-of-the-art stereo methods on the SceneFlow in-domain setting, as shown in Table~\ref{tab:sceneflow}. Among methods without any monocular VFM, WAVE-Stereo achieves the highest accuracy with EPE 0.43. Notably, WAVE-Stereo not only comprehensively outperforms all real-time non-VFM methods---reducing error by 14.0\% over LightStereo-H~\cite{lightstereo} (EPE 0.50), 14.0\% over RT-IGEV~\cite{igevpp} (EPE 0.50), and 29.5\% over RAFT-Stereo~\cite{raft-stereo} (EPE 0.61)---but also surpasses several non-real-time non-VFM methods: 10.4\% error reduction over IGEV-Stereo~\cite{igev} (EPE 0.48), 6.5\% over IGEV++~\cite{igevpp} (EPE 0.46), and 2.3\% over Selective-IGEV~\cite{selectiveigev} (EPE 0.44). Among these, IGEV++ and Selective-IGEV both rely on computationally expensive 3D convolution cost aggregation and multi-level geometry encoding volumes, yet WAVE-Stereo achieves competitive in-domain accuracy through GWCE's unified geometric-warping encoding and PGCP's global context injection, without any external priors. The accuracy gap to VFM methods remains within a manageable range (see \S\ref{sec:zeroshot} for efficiency comparison).
\vspace{0.15in}

\subsection{Ablation Studies}
\label{sec:ablation}

We conduct four groups of ablation experiments to systematically verify the contribution and design choices of each module in WAVE-Stereo: component-wise build-up verifies the independent marginal contribution of each module, GWCE branch ablation verifies the necessity of unified correspondence modeling, PGCP design choices verify the optimal configuration of global context injection, and runtime decomposition verifies the real-time design objective. All ablation experiments are trained on SceneFlow and evaluated on the test set by EPE.
\subsubsection{Component Ablation}
\label{sec:build_up}

To verify the independent contribution of GWCE and PGCP, we start from a baseline model and progressively add each module. All configurations are evaluated on the SceneFlow test set for accuracy and inference speed.

\input{table/ablation_build_up}

The baseline adopts IGEV~\cite{igev}-style ConvGRU iterative optimization, using only correlation retrieval as the motion encoder. Adding PGCP alone yields a modest improvement (0.57 EPE), verifying that global context provides a marginal benefit even without unified correspondence encoding. Introducing GWCE unified encoding substantially improves accuracy (0.46 EPE), validating the complementary gain of warping and correlation features. Further adding PGCP periodic global context injection achieves the optimal 0.43 EPE, demonstrating that global spatial constraints provide critical supplementary information for local iterative updates.
\subsubsection{GWCE Branch Ablation}
\label{sec:gwce_ablation}

To evaluate the contribution of each GWCE branch, we compare three configurations: (A) w/ Correlation, corresponding to IGEV~\cite{igev}'s matching search paradigm; (B) w/ Warping, corresponding to WAFT~\cite{waft-stereo}'s feature warping paradigm; and (C) Full, unifying both. All configurations retain the disparity prior branch for fair comparison.

\input{table/ablation_gwce}

As shown in Table~\ref{tab:gwce_abl}, each branch contributes complementary information. w/ Correlation (A) achieves 0.451 EPE, w/ Warping (B) achieves 0.447 EPE. Full (C) achieves 0.434 EPE, reducing error by 2.9\% over (B) and 3.8\% over (A). Notably, w/ Warping (B) slightly outperforms w/ Correlation (A), indicating that within the current iterative framework, residual alignment signals are more critical than matching search signals. However, their joint use produces an effect surpassing any single paradigm, validating our core insight---correlation volume and feature warping are complementary rather than alternative correspondence representations.
\subsubsection{PGCP Structural Design}
\label{sec:pgcp_ablation}

\input{table/ablation_pgcp}

The key principle of PGCP is to supplement ConvGRU's local updates with a sufficiently compact global branch, rather than stacking a heavy Transformer. We therefore compress PGCP's hyperparameters into a small set of representative configurations following the exploration paradigm commonly used in stereo matching literature. All configurations fix the periodic injection strategy ($K=3$) and vary only the hidden dimension, number of Transformer layers, attention heads, and MLP expansion ratio to analyze how PGCP's internal structural capacity affects accuracy and speed.

\begin{figure}[ht]
\centering
\includegraphics[width=\columnwidth]{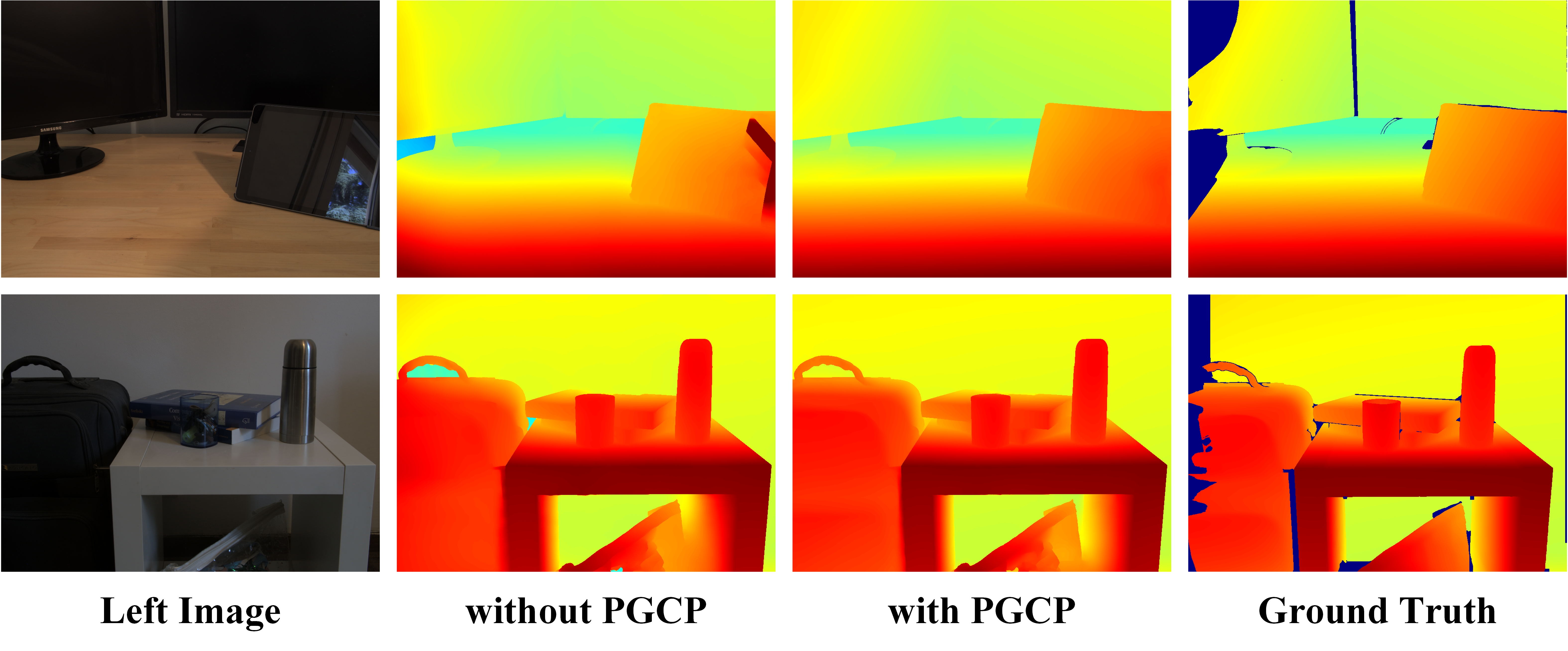}
\caption{Ablation visualization of the PGCP module. From left to right: RGB images, without PGCP, with PGCP, and ground truth. PGCP effectively reduces local estimation errors and improves disparity continuity in textureless regions.}
\label{fig:pgcp_compare}
\vspace{-10pt}
\end{figure}

As shown in Table~\ref{tab:pgcp_abl}, config b (128-dim hidden space, 3 Transformer layers) achieves the optimal 0.43 EPE, outperforming both config a (0.46 EPE) and config c (0.45 EPE), indicating that the global branch requires moderate capacity to effectively model long-range spatial dependencies: too small a capacity (a) cannot propagate global information, while simply enlarging the hidden dim and head count (c) yields limited gains and raises PGCP time from 3.2ms to 4.5ms. Config d increases the layer count to 4 and the MLP ratio to 4, yet EPE degrades to 0.49, showing that blindly deepening or widening the Transformer introduces optimization difficulty rather than performance gains. Config e switches the injection interval of b from $K{=}3$ to every-iteration injection ($K{=}1$), which raises the cost to 8.8ms but degrades EPE to 0.48, confirming that overly frequent injection not only adds overhead but may also perturb the local iterative update by repeatedly injecting redundant global signals. Overall, config b achieves the best accuracy-efficiency balance and is thus selected as the final configuration. Fig.~\ref{fig:pgcp_compare} provides an ablation visualization of the improvement brought by PGCP on depth estimation quality.

\subsubsection{Runtime Decomposition}
\label{sec:runtime}

\input{table/ablation_runtime}

WAVE-Stereo is designed for real-time inference. To verify whether the efficiency design of each component meets expectations, we measure the runtime breakdown at 8 iterations. The full pipeline (including CPU pre/post-processing) is approximately 65ms per frame.

As shown in Table~\ref{tab:abl_runtime}, the GPU forward pass totals 60.32ms, meeting the real-time requirement. The runtime distribution clearly reflects WAVE-Stereo's architectural philosophy of ``lightweight front-end + strong iterative refinement'': front-end processing (feature extraction, cost volume, initial disparity) accounts for 32.2\%, while the computational budget is concentrated in the iterative refinement stage (GWCE + ConvGRU collectively 57.4\%). The efficiency of the two core modules aligns with design expectations---GWCE executes all three branches at every iteration yet costs only 2.32ms per call due to exclusive use of lightweight convolutions; PGCP operates at 1/32 resolution and is invoked only once every 3 iterations, accounting for merely 4.8\%, demonstrating that periodic injection introduces global context with low computational overhead.

\subsection{Limitations}
\label{sec:qual}

(1) Although our method demonstrates strong generalization, the model is primarily trained on synthetic data. Incorporating real-world pseudo-label data during training may further improve performance. (2) The maximum disparity is limited to 192 pixels. This is often insufficient for close-range scenes with large baselines and high resolutions. 

%% file: table/zero_shot.tex
\begin{table*}[t]
\centering
\footnotesize
\setlength{\tabcolsep}{4.5pt}
\vspace{-8pt}
\caption{Zero-shot generalization comparison on five real-world benchmarks. \textbf{Bold}: Best.}
\resizebox{\textwidth}{!}{%
\begin{tabular}{lcccccccc}
\toprule
Method & KITTI-15 D1 $\downarrow$ & KITTI-12 D1 $\downarrow$ & ETH3D BP-2 $\downarrow$ & Booster BP-2 $\downarrow$ & Booster BP-3 $\downarrow$ & Middlebury BP-3 $\downarrow$ & Speed (ms) & Memory (MB) \\
\midrule
\rowcolor[rgb]{ .906, .902, .902} IGEV++~\cite{igevpp} & 9.92 & 5.10 & 2.29 & 16.99 & 14.42 & 7.77 & 582 & 2536 \\
MonSter~\cite{monster} & 3.39 & 3.01 & 0.54 & 13.00 & 9.51 & 3.49 & 1301 & 6544 \\
\rowcolor[rgb]{ .906, .902, .902} WAFT-Stereo~\cite{waft-stereo} & 3.00 & 2.71 & \textbf{0.32} & 5.35 & 3.61 & 1.91 & 601 & 10412 \\
FoundationStereo~\cite{foundationstereo} & \textbf{2.88} & \textbf{2.48} & 0.38 & \textbf{4.35} & 3.46 & \textbf{1.69} & 1852 & 13142 \\
\midrule
RT-IGEV~\cite{igevpp} & 6.65 & 6.18 & 3.10 & 23.20 & 19.00 & 11.11 & 79 & 1070 \\
\rowcolor[rgb]{ .906, .902, .902} LiteAnyStereo~\cite{liteanystereo} & 3.72 & 2.97 & 1.67 & 17.07 & 12.89 & 7.05 & 43 & 902 \\
MatchStereo-Tiny~\cite{yan2025matchattention} & 3.93 & 2.87 & 0.93 & 7.94 & 6.29 & 4.30 & 82 & 920 \\
\rowcolor[rgb]{ .906, .902, .902} Fast-FoundationStereo~\cite{fastfoundationstereo} & 3.21 & \textbf{2.54} & \textbf{0.52} & 7.15 & 5.75 & \textbf{2.63} & 166 & 2732 \\
\textbf{WAVE-Stereo (Ours)} & \textbf{3.18} & 3.06 & 0.86 & \textbf{4.42} & \textbf{2.84} & 3.92 & 66 & 980 \\
\bottomrule
\end{tabular}%
}
\label{tab:zeroshot}
\vspace{-10pt}
\end{table*}

%% file: table/sceneflow.tex
\begin{table}[t]
\centering
\vspace{-0.1in}
\caption{SceneFlow test set in-domain accuracy comparison. WAVE-Stereo achieves the best EPE among methods without monocular VFM; VFM methods serve as accuracy upper bounds.}
\label{tab:sceneflow}
\footnotesize
\begin{tabular}{lc}
\toprule
Method & EPE $\downarrow$ \\
\midrule
\multicolumn{2}{c}{\textit{Without Monocular VFM}} \\
\midrule
LightStereo-H~\cite{lightstereo} & 0.50 \\
LiteAnyStereo~\cite{liteanystereo} & 0.82 \\
RAFT-Stereo~\cite{raft-stereo} & 0.61 \\
IGEV-Stereo~\cite{igev} & 0.48 \\
RT-IGEV~\cite{igevpp} & 0.50 \\
IGEV++~\cite{igevpp} & 0.46 \\
Selective-IGEV~\cite{selectiveigev} & 0.44 \\
\textbf{WAVE-Stereo (Ours)}  &  \textbf{0.43} \\
\midrule
\multicolumn{2}{c}{\textit{With Monocular VFM (Upper Bound)}} \\
\midrule
MonSter~\cite{monster} & 0.37 \\
FoundationStereo~\cite{foundationstereo} & 0.34 \\
\bottomrule
\end{tabular}
\vspace{-15pt}
\end{table}

%% file: table/ablation_build_up.tex
\begin{table}[ht]
\centering
\def\mywidth{0.32\textwidth}
\definecolor{green}{RGB}{0,200,0}
\renewcommand{\arraystretch}{0.9}
\vspace{-0.1in}
\caption{Component-wise build-up ablation on SceneFlow test set. Row~1 is the IGEV-style baseline. The full model is highlighted in green.}
\resizebox{\mywidth}{!}{%

\begin{tabular}{cccc}
\cmidrule{1-4}
Row & GWCE & PGCP & EPE $\downarrow$ \\
\cmidrule{1-4}
1 & & & 0.58 \\
2 & & \checkmark & 0.57 \\
3 & \checkmark & & 0.46 \\
4 & \cellcolor[rgb]{ .886, .937, .855}\checkmark & \cellcolor[rgb]{ .886, .937, .855}\checkmark & \cellcolor[rgb]{ .886, .937, .855}\textbf{0.43} \\
\cmidrule{1-4}
\end{tabular}%

}
\label{tab:abl_build}
\vspace{-15pt}
\end{table}

%% file: table/ablation_gwce.tex
\begin{table}[ht]
\centering
\def\mywidth{0.42\textwidth}
\definecolor{green}{RGB}{0,200,0}
\vspace{-0.1in}
\caption{GWCE branch ablation on SceneFlow test set. All configurations retain the disparity prior branch for fair comparison. The full model is highlighted in green.}
\resizebox{\mywidth}{!}{%

\begin{tabular}{lccc}
\cmidrule{1-4}
Configuration & Correlation & Warping & EPE $\downarrow$ \\
\cmidrule{1-4}
A: w/ Correlation & \checkmark & & 0.451 \\
B: w/ Warping & & \checkmark & 0.447 \\
\textbf{C: Full (Ours)} & \cellcolor[rgb]{ .886, .937, .855}\checkmark & \cellcolor[rgb]{ .886, .937, .855}\checkmark & \cellcolor[rgb]{ .886, .937, .855}\textbf{0.434} \\
\cmidrule{1-4}
\end{tabular}%

}
\label{tab:gwce_abl}
\vspace{-15pt}
\end{table}

%% file: table/ablation_pgcp.tex
\begin{table*}[t]
\centering
\def\mywidth{0.85\textwidth}
\definecolor{green}{RGB}{0,200,0}
\vspace{-0.1in}
\caption{PGCP structural design ablation on SceneFlow test set. Configurations a--d use periodic injection ($K=3$); e uses the same hyperparameters as b but with every-iteration injection ($K=1$), validating that periodic injection achieves similar accuracy at a fraction of the cost. The default configuration (b) is highlighted in green. PGCP time measured with the same setup as Table~\ref{tab:abl_runtime}.}
\resizebox{\mywidth}{!}{%

\begin{tabular}{lcccccc}
\cmidrule{1-7}
& TF\_DIM & TF\_LAYERS & TF\_HEADS & TF\_MLP\_RATIO & EPE $\downarrow$ & PGCP Time (ms) \\
\cmidrule{1-7}
a & 64 & 2 & 4 & 2 & 0.46 & 1.9 \\
b & \cellcolor[rgb]{ .886, .937, .855}128 & \cellcolor[rgb]{ .886, .937, .855}3 & \cellcolor[rgb]{ .886, .937, .855}4 & \cellcolor[rgb]{ .886, .937, .855}3 & \cellcolor[rgb]{ .886, .937, .855}0.43 & \cellcolor[rgb]{ .886, .937, .855}3.2 \\
c & 256 & 3 & 8 & 3 & 0.45 & 4.5 \\
d & 128 & 4 & 4 & 4 & 0.49 & 3.3 \\
e & 128 & 3 & 4 & 3 & 0.48 & 8.8 \\
\cmidrule{1-7}
\end{tabular}%

}
\label{tab:pgcp_abl}
\vspace{8pt}
\end{table*}

%% file: table/ablation_runtime.tex
\begin{table}[t]
\centering
\def\mywidth{0.48\textwidth}
\vspace{-0.1in}
\caption{Runtime decomposition at 8 iterations. The full pipeline (including CPU pre/post-processing) is approximately 65ms per frame.}
\resizebox{\mywidth}{!}{%

\begin{tabular}{llcc}
\cmidrule{1-4}
Stage & Component & Time (ms) & Proportion (\%) \\
\cmidrule{1-4}
\multirow{3}*{Frontend ($\times$1)} & Feature Extraction (MobileNetV2) & 10.72 & 17.8 \\
& Cost Volume + 2D Aggregation & 8.58 & 14.2 \\
& Initial Disparity Regression & 0.15 & 0.3 \\
& \textit{Subtotal} & \textit{19.45} & \textit{32.2} \\
\cmidrule{1-4}
\multirow{2}*{Iterative ($\times$8)} & GWCE Three-Branch Encoding & 18.53 & 30.7 \\
& ConvGRU Update & 16.11 & 26.7 \\
& \textit{Subtotal} & \textit{34.64} & \textit{57.4} \\
\cmidrule{1-4}
Periodic ($\times$3, $K=3$) & PGCP Global Context Injection & 2.92 & 4.8 \\
\cmidrule{1-4}
Backend ($\times$1) & Disparity Upsampling & 3.32 & 5.5 \\
\cmidrule{1-4}
\textbf{Total} & & \textbf{60.32} & \textbf{100} \\
\cmidrule{1-4}
\end{tabular}%

}
\label{tab:abl_runtime}
\vspace{-15pt}
\end{table}

%% file: sec/5_conclusion.tex
\section{Conclusion}
\label{sec:conclusion}

We proposed WAVE-Stereo, an iterative optimization framework for real-time zero-shot stereo matching. To address the limitation of existing methods that adopt either correlation search or feature warping in isolation, we introduce the GeoWarp Correspondence Encoder (GWCE), which jointly models matching candidate information, cross-view alignment information, and geometric priors, and design Periodic Global Context Propagation (PGCP) to periodically inject global spatial constraints with low overhead. Experiments show that WAVE-Stereo achieves competitive zero-shot generalization and real-time inference speed on multiple real-world datasets without relying on any external monocular depth foundation model. Future work will further explore unified correspondence modeling and leverage larger-scale training data to improve generalization in complex scenes.